\documentclass[10pt]{article}
\usepackage{natbib}
\usepackage{times}
\usepackage[small,normal,bf,up]{caption2}
\makeatletter
\def\@seccntformat#1{\normalsize{\csname the#1\endcsname.\quad}}
\makeatother

\usepackage[final]{graphicx} % need this to include pictures
\usepackage{url} % needed to display URLs properly
\usepackage{ams} % some symbols which I require
\usepackage{algorithmic} % this is great to display pseudo-code
\usepackage{algorithm} % goes together with the previous package
\newtheorem{definition}{Definition} % new theorem for definitions
\begin{document}

\begin{center} MCDM 2006, Chania, Greece, June 19-23, 2006

\setlength{\parskip}{25pt}\large{\bf{FOUNDATIONS OF THE PARETO
ITERATED LOCAL SEARCH METAHEURISTIC}}
\end{center}

\normalsize

\begin{center}

\vspace{15pt}

\large\textbf{Martin Josef Geiger}\normalsize

Lehrstuhl f\"{u}r Industriebetriebslehre

Institut f\"{u}r Betriebswirtschaftslehre

Universit\"{a}t Hohenheim

Schlo\ss{} Hohenheim, Osthof-Nord, D-70593 Stuttgart, Germany

E-mail: mjgeiger@uni-hohenheim.de
\end{center}

\vspace{15pt}\noindent \textbf{Keywords}: pareto iterated local
search, metaheuristics, multi-objective scheduling

\vspace{10pt}\hspace{1.25cm}\parbox{13.75cm}{\textbf{Summary:}
\textit{The paper describes the proposition and application of a
local search metaheuristic for multi-objective optimization
problems. It is based on two main principles of heuristic search,
intensification through variable neighborhoods, and diversification
through perturbations and successive iterations in favorable regions
of the search space. The concept is successfully tested on
permutation flow shop
scheduling problems under multiple objectives. %, significantly outperforming know approaches from evolutionary computation.
While the obtained results are encouraging in terms of their
quality, another positive attribute of the approach is its'
simplicity as it does require the setting of only very few
parameters.\\The implementation of the Pareto Iterated Local Search
metaheuristic is based on the MOOPPS computer system of local search
heuristics for multi-objective scheduling which has been awarded the
European Academic Software Award 2002 in Ronneby, Sweden
(\url{http://www.easa-award.net/},
\url{http://www.bth.se/llab/easa_2002.nsf)}.}}

\setlength{\parskip}{10pt}

\section{\normalsize Introduction}\vspace{-10pt}
Real world problems often comprise several points of view that from
a decision makers perspective have to be taken simultaneously into
consideration. Multi-objective optimization approaches play in this
context an increasingly important role, tackling applications in
numerous areas. Due to the complexity of most problems however,
problem resolution has to rely in many cases on modern heuristics
that provide fast results without necessarily identifying an optimal
solution. Here, local search approaches like e.\,g.\ Simulated
Annealing, Evolutionary Algorithms, and Tabu Search play a dominant
role. Depending on the application area, more and more refined
version and adaptations of local search metaheuristics have been
proposed with increasing success in recent years.

Scheduling is one of the most active areas of research, with
applications in numerous areas of manufacturing, computer
systems/grid scheduling, sports/tournament scheduling, and
airline/fleet scheduling, to mention a few. Many of the mentioned
problems are of multi-criteria nature, and considerable effort has
been made to solve these often $\mathcal{NP}$-hard problems. While
metaheuristics often lead to acceptable results, room for
improvements can still be identified, especially as modern
metaheuristics tend to require increasingly complex parameter
settings.

The current paper describes an local search heuristic for the
effective resolution of multi-objective optimization problems, based
on the local search paradigm. An application of the approach is
presented to the multi-objective permutation flow shop scheduling
problem. The article is organized as follows.
Section~\ref{sec:mo:scheduling} first introduces the considered
problem and briefly reviews heuristic solution approaches known from
literature. The Pareto Iterated Local Search algorithm is then
presented in Section~\ref{sec:pils}. An application of the
metaheuristic to the discussed problem is given in the following
Section~\ref{sec:application}, and conclusions are drawn in
Section~\ref{sec:conclusions}.

\section{\normalsize \label{sec:mo:scheduling}Solving the Multi-Objective Permutation Flow Shop Scheduling Problem by Metaheuristics}%
\vspace{-10pt}\subsection{\normalsize Problem Description}\vspace{-10pt} %
The flow shop scheduling problem consists in the assignment of a set
of jobs $\mathcal{J} = \{ J_{1}, \ldots J_{n} \}$, each of which
consists of a set of operations $J_{j} = \{ O_{j1}, \ldots,
O_{jo_{j}} \}$ onto a set of machines $\mathcal{M} = \{ M_{1},
\ldots, M_{m} \}$ (B{\l}a{\.{z}}ewicz, Ecker, Pesch, Schmidt and
W\c{e}glarz, 2001; Pinedo, 2002). Each operation $O_{jk}$ is
processed by at most one machine at a time, involving a non-negative
processing time $p_{jk}$. The result of the problem resolution is a
schedule $x$, defining for each operation $O_{jk}$ a starting time
$s_{jk}$ on the corresponding machine. Several side constraints are
present which have to be respected by any solution $x$ belonging to
the set of feasible schedules $X$. Precedence constraints $O_{jk}
\rhd O_{jk+1} \forall j = 1, \ldots, n, k = 1, \ldots, o_{j}-1$
between the operations of a job $J_{j}$ assure that processing of
$O_{jk+1}$ only commences after completion of $O_{jk}$, thus
$s_{jk+1} \geq s_{jk} + p_{jk}$. In flow shop scheduling, the
machine sequence in which the operations are processed by the
machines is identical for all jobs, and for the specific case of the
permutation flow shop scheduling the job sequence must also be the
same on all machines.

The assignment of operations to machines has to be done with respect
to one or several optimality criteria. Most optimality criteria are
functions of the completion times $C_{j}$ of the jobs $J_{j}$, with
the computation as given in Expression~(\ref{eqn:completion:time}).
\begin{equation}%
\label{eqn:completion:time} C_{j} = s_{jo_{j}} + p_{jo_{j}}%
\end{equation}%
The most prominent is the maximum completion time (makespan)
$C_{max}$, computed in the following
Expression~(\ref{eqn:makespan}).
\begin{equation}%
\label{eqn:makespan} C_{max} = \max C_{j}%
\end{equation}%
Others express violations of due dates $d_{j}$ of jobs $J_{j}$. A
due date $d_{j}$ defines a latest point of time until a job $J_{j}$
should be finished as the assembled product has to be delivered to
the customer on this date. The computation of an occurring tardiness
$T_{j}$ of a job $J_{j}$ is given in
Expression~(\ref{eqn:tardiness}). A possible optimality criteria
based on tardiness of jobs is e.\,g.\ the total tardiness $T_{sum}$
as given in Expression (\ref{eqn:total:tardiness}).
\begin{eqnarray}%
\label{eqn:tardiness} T_{j} & = & \max (C_{j} - d_{j}, 0)\\%
\label{eqn:total:tardiness} T_{sum} & = & \sum T_{j}%
\end{eqnarray}%
It is known, that for \emph{regular} optimality criteria at least
one \emph{active} schedule $x$ does exist which is also optimal
(Baker, 1974). %Consequently,
%many approaches restrict their search to the set of active
%schedules.
As the representation of an active schedule for the permutation flow
shop scheduling problem is possible using a permutation of jobs $\pi
= ( \pi_{1}, \ldots, \pi_{n} )$, where each $\pi_{j}$ stores a job
$J_{k}$ at position $j$, this way of representing alternatives is
often used in resolution approaches. The search is then restricted
to the much smaller set of active schedules only.

Multi-objective approaches to scheduling consider a vector $G(x) =
(g_{1}(x), \ldots, g_{K}(x))$ of optimality criteria at once
(T'kindt and Billaut, 2002). As the relevant optimality criteria are
often of conflicting nature, not a single solution $x \in X$ exists
optimizing all components of $G(x)$ at once. Optimality in
multi-objective optimization problems is therefore understood in the
sense of Pareto-optimality, and the resolution of multi-objective
optimization problems lies in the identification of all elements
belonging to the Pareto set $P$, containing all alternatives $x$
which are not dominated by any other alternative $x' \in X$. The
corresponding definitions are given in
Definition~\ref{def:dominance} and \ref{def:pareto:optimality}.
Without loss of generality we assume the minimization of the
optimality criteria $g_{i}(x) \forall i = 1, \ldots, K$.
\begin{definition}[Dominance]%
\label{def:dominance}A vector $G(x), x \in X$ is said to dominate a
vector $G(x'), x' \in X$ if and only if $g_{i}(x) \leq g_{i}(x')
\wedge \exists i \mid g_{i}(x) < g_{i}(x')$. We denote the dominance
of a $G(x)$ over $G(x')$ with $G(x) \preceq G(x')$.
\end{definition}%
\begin{definition}[Pareto-optimality, Pareto set]%
\label{def:pareto:optimality}An alternative $x, x \in X$ is called
Pareto-optimal if and only if $\not\!\exists x' \in X \mid G(x')
\preceq G(x)$. The corresponding vector $G(x)$ of a Pareto-optimal
alternative is called efficient, the set of all Pareto-optimal
alternatives is called the Pareto set $P$.
\end{definition}
After the identification of the Pareto set $P$, an interactive
search might be performed by the decision maker (Vincke, 1992). The
interactive procedure terminates with the identification of a
most-preferred solution $x^{*} \in P$.

\vspace{-10pt}\subsection{\normalsize Previous Research}\vspace{-10pt} %
Several approaches of metaheuristics have been formulated and tested
in order to solve the permutation flow shop scheduling problem under
multiple, in most cases two, objectives. Common to all is the
representation of solutions using permutations $\pi$ of jobs, as in
previous investigation only regular functions are considered.

First results have been obtained using Evolutionary Algorithms,
which in general play a dominant role in the resolution of
multi-objective optimization problems when using metaheuristics.
This is mainly due to the fact that these methods incorporate the
idea of a set of solutions, a so called \emph{population}, as a
general ingredient. Flow shop scheduling problems minimizing the
maximum completion time and the average flow time have been solved
by (Nagar, Heragu and Haddock, 1996). In their work, they however
combine the two objectives into a weighted sum. Problems minimizing
the maximum completion time and the total tardiness are solved by
(Murata, Ishibuchi and Tanaka, 1996), again under the combination of
both objectives into a weighted sum. Later work on the same problem
class by (Basseur, Seynhaeve and Talbi, 2002) avoids the weighted
sum approach, using dominance relations among the solutions
only.\\Most recent work is presented by (Loukil, Teghem and
Tuyttens, 2005). Contrary to approaches from Evolutionary
Computations, the authors apply the Multi Objective Simulated
Annealing approach MOSA (Ulungu, Teghem, Fortemps and Tuyttens,
1999) to a variety of bi-criterion scheduling problems.

Flow shop scheduling problems with three objectives are studied by
(Ishibuchi and Murata, 1998), and (Ishibuchi, Yoshida and Murata,
2003). The authors minimize the maximum completion time, the total
completion time, and the maximum tardiness at once. A similar
problem minimizing the maximum completion time, the average flow
time, and the average tardiness is then tackled by (Bagchi, 1999;
Bagchi, 2001).

\section{\normalsize \label{sec:pils}Pareto Iterated Local Search}\vspace{-10pt} %
The Pareto Iterated Local Search (PILS) metaheuristic is a novel
concept for the resolution of multi-objective optimization problems.
It combines the two main driving forces of local search,
intensification and diversification, into a single algorithm. The
motivation behind the proposition of this concept can be seen in the
increasing demand for simple, yet effective heuristics for the
resolution of complex multi-objective optimization problems. Two
developments in local search demonstrate the effectiveness of some
intelligent ideas that make use of certain structures within the
search space topology of problems. First, Iterated Local Search
(Louren\c{c}o, Martin and St\"{u}tzle, 2003), introducing the idea
of perturbations to overcome local optimality and continue search in
interesting areas of the search space. Second, Variable Neighborhood
Search (Hansen and Mladenovi\'{c}, 2003), combining multiple
neighborhood operators into a single algorithm in order to avoid
local optimality in the first place. In the proposed concept, both
paradigms are combined and extended within a search framework
handling not only a single but a set of alternatives at once.

\begin{figure}[!ht]
\begin{center}
\includegraphics{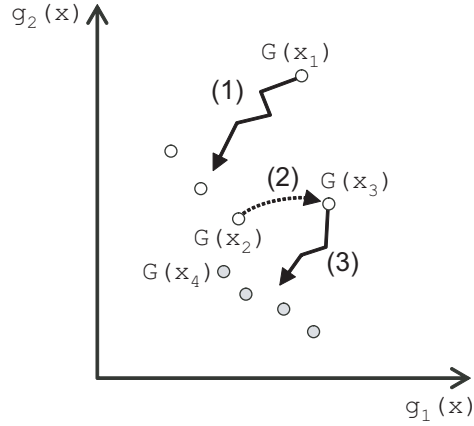}
\end{center}
\caption{\label{fig:pils}Illustration of the Pareto Iterated Local
Search metaheuristic. The archive of the currently best solutions is
updated during the search. Here, $G(x_{4})$ dominates $G(x_{2})$
which is going to be deleted from $P^{approx}$.}
\end{figure}

The main principle of the algorithm is sketched in Figure
\ref{fig:pils}. Starting from an initial solution $x_{1}$, an
improving, intensifying search is performed until a set of locally
optimal alternatives is identified, stored in a set $P^{approx}$
representing the approximation of the true Pareto set $P$. No
further improvements are possible from this point. In this initial
step, a set of neighborhoods ensures that all identified
alternatives are locally optimal not only to a single but to a set
of neighborhoods. This principle, known from Variable Neighborhood
Search, promises to lead to better results as it is known that all
global optima are also locally optimal with respect to all possible
neighborhoods while this is not necessarily the case for local
optima.

After the identification of a locally optimal set, a diversification
step is performed on a solution $x_{2}$ using a perturbation
operator, continuing search from the perturbed solution $x_{3}$. The
perturbation operator has to be significantly different from the
neighborhoods used in intensification, as otherwise the following
search would return to the previous solution. On the other hand
however, the perturbation should not entirely destroy the
characteristics of the alternative. Doing that would result in a
random restart of the search without keeping promising attributes of
solutions.

The PILS metaheuristic may be formalized as given in
Algorithm~\ref{alg:pils}. The intensification of the algorithm,
illustrated in the steps (1) and (3) of Figure~\ref{fig:pils} is
within the lines 6 to 21, the description of the diversification,
given in step (2) of Figure~\ref{fig:pils} is within the lines 22 to
26.

\begin{algorithm}[!ht]
\caption{\label{alg:pils}Pareto Iterated Local Search}
\begin{algorithmic}[1]
\STATE{Initialize control parameters: Define the neighborhoods $\mathbf{N}_{1}, \ldots, \mathbf{N}_{k}$}%
\STATE{Set $i = 1$} %
\STATE{Generate initial solution $x$}%
\STATE{Set $P^{approx} = \{ x \}$}%
\REPEAT%
    \REPEAT%
        \STATE{Compute $\mathbf{N}_{i}(x)$}%
        \STATE{Evaluate $\mathbf{N}_{i}(x)$}%
        \STATE{Update $P^{approx}$ with $\mathbf{N}_{i}(x)$}%
        \IF{$\exists x' \in \mathbf{N}_{i}(x) \mid x' \preceq x$}%
            \STATE{Set $x = x'$}%
            \STATE{Set $i = 1$}%
            \STATE{Rearrange the neighborhoods $\mathbf{N}_{1}, \ldots, \mathbf{N}_{k}$ in some random order}%
        \ELSE%
            \STATE{Set $i = i + 1$}%
        \ENDIF%
    \UNTIL{$x$ locally optimal with respect to $\mathbf{N}_{1}, \ldots, \mathbf{N}_{k}$, therefore $i > k$}%
    \STATE{Set neighborhoods of $x$ as \lq{}investigated\rq{}}
    \STATE{Set $i = 1$}
    \IF{$\exists x' \in P^{approx} \mid$ neighborhoods not investigated yet}%
        \STATE{Set $x = x'$}%
    \ELSE%
        \STATE{Select $x' \in P^{approx}$}%
        \STATE{Compute $x'' = \mathbf{N}_{perturb}(x')$}%
        \STATE{Set $x = x''$}%
    \ENDIF%
\UNTIL{termination criterion is met}%
\end{algorithmic}
\end{algorithm}

It can be seen, that the algorithm computes a set of neighborhoods
for each alternative. The sequence in which the neighborhoods are
computed is arranged in a random fashion, described in line 13 of
Algorithm~\ref{alg:pils}. This introduces an additional element of
diversity to the concept, as otherwise the search might be biased by
a certain sequence of neighborhoods.

\section{\normalsize \label{sec:application}An Application to Multi-Objective Flow Shop Scheduling} %
\vspace{-10pt}\subsection{\normalsize Configuration of the Algorithm and Experimental Setup}\vspace{-10pt} %
In the following, the Pareto Iterated Local Search is applied to a
set of benchmark instances of the multi-objective permutation flow
shop scheduling problem. They have been provided by (Basseur,
Seynhaeve and Talbi, 2002), who first defined due dates for the
well-known instances of (Taillard, 1993). The instances range from
$n = 20$ jobs that have to be processed on $m = 5$ machines to $n =
100, m = 20$. All of them are solved under the simultaneous
consideration of the minimization of the maximum completion time
$C_{max}$ and the total tardiness $T_{sum}$.

Three operators are used in the definition of the neighborhoods
$\mathbf{N}_{1}, \ldots, \mathbf{N}_{k}$, described in the work of
(Reeves, 1999). First, an exchange neighborhood, exchanging the
position of two jobs in $\pi$, second, a forward shift neighborhood,
taking a job from position $i$ and reinserting it at position $j$
with $j < i$, and finally a backward shift neighborhood, shifting a
job from position $i$ to $j$ with $j < i$. All operators are problem
independent, each computing $\frac{n(n-1)}{2}$ neighboring
solutions.

After a first approximation $P^{approx}$ of the Pareto set is
obtained, one element $x' \ P^{approx}$ is selected by random and
perturbed into another solution $x''$. We use a special neighborhood
that on one hand leaves most of the characteristics of the perturbed
alternatives intact, while still changes the positions of some jobs.
Also, several consecutive applications of the neighborhoods
$\mathbf{N}_{1}, \ldots, \mathbf{N}_{k}$ are needed to return from
$x''$ back to $x'$. This is important, als otherwise the algorithm
might navigate straight back to the initially perturbed alternative,
possibly leading to a cycle in the search path.\\The perturbation
neighborhood $\mathbf{N}_{perturb}$ can be described as follows.
First, a subset of $\pi$ is randomly selected, comprising four
consecutive jobs at positions $j, j+1, j+2, j+3$. Then a neighboring
solution is generated by moving the job at position $j$ to $j+3$,
the one at position $j+1$ to $j+2$, the one at position $j+3$ to
$j$, and the job at position $j+3$ to $j+2$. In brief, this leads to
a combination of several exchange and shift moves, executed at once.
%The jobs at the
%positions $<j$ and $>j+3$ remain however untouched.

The benchmark instances of Basseur have been solved using the PILS
algorithm. In each of the 100 test runs, the approximation quality
of the obtained results has been analyzed using the $D_{1}$ and
$D_{2}$ metrics of (Czy\.{z}ak and Jaszkiewicz, 1998). While for the
smaller instances the optimal solutions are known, the analysis for
the larger instances has to rely on the best known results published
in the literature. Experiments have been carried out on a Intel
Pentium IV processor, running at 1.8 GHz. Table
\ref{tbl:termination:criterion} gives an overview about the number
of evaluations executed for each instance. Clearly, considerable
more alternatives have to be evaluated with increasing size of the
problem instances to allow a convergence of the algorithm.

\begin{table}[!ht]
\begin{center}
\caption{\label{tbl:termination:criterion}Number of evaluations for
each investigated instance}
\begin{tabular}{lr}\\
\hline %
Instance $n \times m$  & No of evaluations\\
\hline %
$20 \times 5$ (\#1)  & 1,000,000\\
$20 \times 5$ (\#2)  & 1,000,000\\
$20 \times 10$ (\#1) & 1,000,000\\
$20 \times 10$ (\#2) & 1,000,000\\
$20 \times 20$     & 1,000,000\\
$50 \times 5$      & 5,000,000\\
$50 \times 10$     & 5,000,000\\
$50 \times 20$     & 5,000,000\\
$100 \times 10$    & 10,000,000\\
$100 \times 20$    & 10,000,000\\
\hline%
\end{tabular}
\end{center}
\end{table}

An implementation of the algorithm has been made available within
the MOOPPS computer system, a software for the resolution of
multi-objective scheduling problems using metaheuristics. The system
is equipped with an extensive user interface that allows an
interaction with a decision maker and is able to visualize the
obtained results in alternative and outcome space. The system also
allows the comparison of results obtained by different
metaheuristics. For a first analysis, we compare the results
obtained by PILS to the approximations of a multi-objective
multi-operator search algorithm MOS, described in Algorithm
\ref{alg:mos}.

\begin{algorithm}[!ht]
\caption{\label{alg:mos}Multi-objective multi-operator search
algorithm}
\begin{algorithmic}[1]
\STATE{Generate initial solution $x$, set $P^{approx} = \{ x \}$}%
\REPEAT%
    \STATE{Randomly select some $x \in P^{approx} \mid $ neighborhoods not investigated yet}%
    \STATE{Randomly select some neighborhood $\mathbf{N}_{i}$ from $\mathbf{N}_{1}, \ldots, \mathbf{N}_{k}$}%
    \STATE{Generate $\mathbf{N}_{i}(x)$}%
    \STATE{Update $P^{approx}$ with $\mathbf{N}_{i}(x)$}%
    \IF{$x \in P^{approx}$}%
        \STATE{Set neighborhoods of $x$ as \lq{}investigated\rq{}}%
    \ENDIF%
\UNTIL{$\not\!\exists x \in P^{approx} \mid$ neighborhoods not investigated yet}%
\end{algorithmic}
\end{algorithm}

The MOS Algorithm is based on the concept of Variable Neighborhood
Search, extending the general idea of several neighborhood operators
by adding an archive $P^{approx}$ towards the optimization of
multi-objective problems. For a fair comparison, the same
neighborhood operators are used as in the PILS algorithm. After the
termination criterion is met in step 10, we restart search while
keeping the approximation $P^{approx}$ for the final analysis of the
quality of the obtained solutions.

\vspace{-10pt}\subsection{\normalsize Results}\vspace{-10pt} %
The average values obtained by the investigated metaheuristics are
given in Table \ref{tbl:results}. It can be seen, that PILS leads
for all investigated problem instances to better results for both
the D$_{1}$ and the $D_{2}$ metric. This general result is
consistent independent from the actual problem instance. For a
single instance, the $20 \times 5$~(\#1), PILS was able to identify
all optimal solutions in all test runs, leading to average values of
$D_{1} = D_{2} = 0.0000$. Apparently, this instance is comparably
easy to solve.

\begin{table}[!ht]
\begin{center}
\caption{\label{tbl:results}Average results of $D_{1}$ and $D_{2}$}
\begin{tabular}{lrrrr}
\hline%
& \multicolumn{2}{c}{$D_{1}$} & \multicolumn{2}{c}{$D_{2}$}\\
Instance $n \times m$ & PILS & MOS & PILS & MOS\\
\hline
$20 \times 5$ (\#1)  & 0.0000 & 0.0323 & 0.0000 & 0.1258\\
$20 \times 5$ (\#2)  & 0.1106 & 0.1372 & 0.3667 & 0.4249\\
$20 \times 10$ (\#1) & 0.0016 & 0.0199 & 0.0146 & 0.0598\\
$20 \times 10$ (\#2) & 0.0011 & 0.0254 & 0.0145 & 0.1078\\
$20 \times 20$     & 0.0088 & 0.0286 & 0.0400 & 0.1215\\
$50 \times 5$      & 0.0069 & 0.0622 & 0.0204 & 0.1119\\
$50 \times 10$     & 0.0227 & 0.3171 & 0.0897 & 0.4658\\
$50 \times 20$     & 0.0191 & 0.3966 & 0.0616 & 0.5609\\
$100 \times 10$    & 0.0698 & 0.3190 & 0.1546 & 0.4183\\
$100 \times 20$    & 0.0013 & 0.2349 & 0.0255 & 0.3814\\
\hline
\end{tabular}
\end{center}
\end{table}

\begin{figure}[!ht]
\begin{center}
\includegraphics{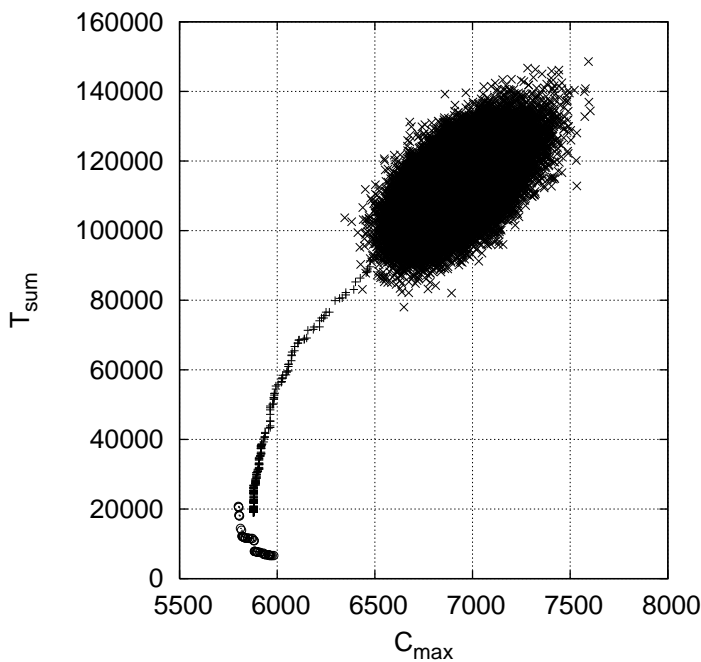}
\end{center}
\caption{\label{fig:rand:intens:front}Randomly generated solutions
($\times$), intensification of search($+$), and Pareto front
($\odot$)}
\end{figure}

A deeper analysis has been performed to monitor the resolution
behavior of the local search algorithms and to get a better
understanding of how the algorithm converges towards the Pareto
front. Figure \ref{fig:rand:intens:front} plots with $\times$ the
results obtained by random sampling 50,000 alternatives for the
problem instance $100 \times 10$, and compares the points obtained
during the first intensification procedure of PILS until a locally
optimal set is identified. The alternatives computed starting from a
random initial solution towards the Pareto front are plotted as $+$,
the Pareto front as $\odot$. It can be seen, that in comparison to
the initial solution even a simple local search approach converges
in rather close proximity to the Pareto front. With increasing
number of computations however, the steps towards the optimal
solutions get increasingly smaller, as it can be seen when
monitoring the distances between the $+$ symbols. After convergence
towards a locally optimal set, overcoming local optimality is then
provided by means of the perturbation neighborhood
$\mathbf{N}_{perturb}$.

\begin{figure}[!ht]
\begin{center}
\includegraphics{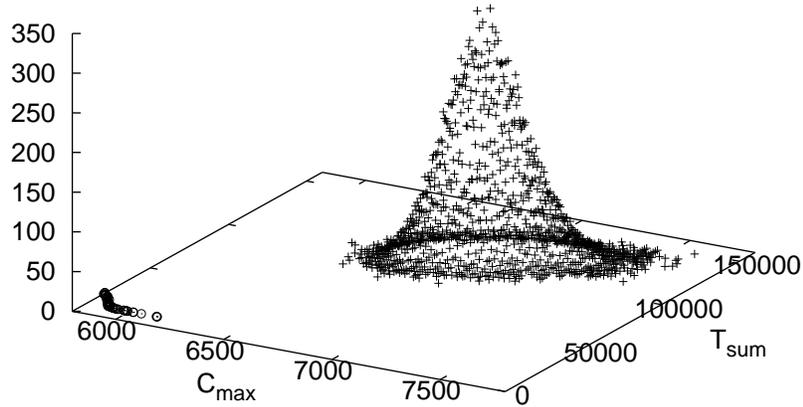}
\end{center}
\caption{\label{fig:dist:random}Distribution of randomly generated
solutions ($+$) compared to the Pareto front $\odot$)}
\end{figure}

An interesting picture is obtained when analyzing the distribution
of the randomly sampled 50,000 alternatives for instance $100 \times
10$. In Figure \ref{fig:dist:random}, the number of alternatives
with a certain combination of objective function values are plotted
and compared to the Pareto front, given in the left corner. It turns
out that many alternatives are concentrated around some value
combination in the area of approximately $C_{max} = 6900$, $T_{sum}
= 111500$, relatively far away from the Pareto front.

When analyzing the convergence of local search heuristics toward the
globally Pareto front as well as towards locally optimal
alternatives, the question arises how many local search steps are
necessary until a locally optimal alternative is identified. From a
different point of view, this problem is discussed in the context of
computational complexity of local search (Johnson, Papadimitriou and
Yannakakis, 1988). It might be worth investigating this behavior in
quantitative terms. Table \ref{tbl:anz:bis:lokal:optimal} gives the
average number of evaluations that have been necessary to reach a
locally optimal alternative from some randomly generated initial
solution. The analysis reveals that the computational effort grows
exponentially with the number of jobs $n$.

\begin{table}[!ht]
\begin{center}
\caption{\label{tbl:anz:bis:lokal:optimal}Number of evaluations
until a locally optimal alternative is reached}
\begin{tabular}{lrr}\\
\hline %
Instance $n \times m$   & No of jobs & No of evaluations\\
\hline %
$20 \times 5$ (\#1)  & 20  & 3.614\\
$20 \times 5$ (\#2)  & 20  & 3.292\\
$20 \times 10$ (\#1) & 20  & 2.548\\
$20 \times 10$ (\#2) & 20  & 2.467\\
$20 \times 20$       & 20  & 2.657\\
$50 \times 5$      & 50  & 53.645\\
$50 \times 10$     & 50  & 55.647\\
$50 \times 20$     & 50  & 38.391\\
$100 \times 10$    & 100 & 793.968\\
$100 \times 20$    & 100 & 479.420\\
\hline%
\end{tabular}
\end{center}
\end{table}

%\clearpage

\section{\normalsize \label{sec:conclusions}Conclusions}\vspace{-10pt} %
In the past years, considerable progress has been made in the
resolution of complex multi-objective optimization problems.
Effective metaheuristics have been developed, providing the
possibility of computing approximations to problems with numerous
objectives and complex side constraints. While many approaches are
of increasingly effectiveness, complex parameter settings are
however required to tune the solution approach to the given problem
at hand.

The algorithm presented in this paper proposed a metaheuristic,
combining two recent principles of local search, Variable
Neighborhood Search and Iterated Local Search. The main motivation
behind the concept is the easy yet effective resolution of
multi-objective optimization problems with an approach using only
few parameters.

After an initial introduction to the problem domain of flow shop
scheduling under multiple objectives, the introduced PILS algorithm
has been applied to a set of scheduling benchmark instances taken
from literature. We have been able to obtain encouraging results,
despite the simplicity of the algorithmic approach. A comparison of
the approximations of the Pareto sets has been given with a
multi-operator local search approach, and as a conclusion PILS was
able to lead to consistently better results.\\ The presented
approach seems to be a promising tool for the effective resolution
of multi-objective optimization problems. After first tests on
problems from the domain of scheduling, the resolution behavior on
problems from other areas might be an interesting direction for
further developments.

\section*{\normalsize Acknowledgements}\vspace{-10pt} %%
The author would like to thank Matthieu Basseur for providing
multi-objective flow shop scheduling problems on the basis of
(Taillard, 1993) and the corresponding currently best known
alternatives under
\url{http://www.lifl.fr/~basseur/benchs_matth.html}, and Zs\'{i}ros
\'{A}kos (University of Szeged), Pedro Caicedo, Luca Di Gaspero
(University of Udine), and Szymon Wilk (Poznan University of
Technology) for providing multilingual versions of the software
MOOPPS.

\section{\normalsize References}\vspace{-10pt} %%

Bagchi, T.P. (1999), {\em Multiobjective scheduling by genetic
algorithms}, Boston, Dordrecht, London: Kluwer Academic Publishers.

Bagchi, T.P. (2001), ``Pareto-optimal solutions for multi-objective
production scheduling
  problems", in Zitzler, E., Deb, K., Thiele, L., Coello
  Coello, C.A. and Corne, D., editors, {\em Evolutionary Multi-Criterion
  Optimization: Proceedings of the First International Conference EMO 2001},
  Berlin, Heidelberg, New York: Springer Verlag, 458--471.

Baker, K.R. (1974), {\em Introduction to Sequencing and Scheduling},
New York, London, Sydney, Toronto: John Wiley \& Sons.

Basseur, M., Seynhaeve, F. and Talbi, E. (2002), ``Design of
multi-objective evolutionary algorithms: Application to the
flow-shop scheduling problem", in {\em Congress on Evolutionary
Computation (CEC'2002)}, Piscataway, NJ: IEEE Service Center,
1151--1156.

B{\l}a{\.{z}}ewicz, J., Ecker, K.H., Pesch, E., Schmidt, G. and
W\c{e}glarz, J. (2001), {\em Scheduling Computer and Manufacturing
Processes}, Berlin, Heidelberg, New York: Springer Verlag.

Czy\.{z}ak, P. and Jaszkiewicz, A. (1998), ``Pareto simulated
annealing - a metaheuristic technique for
  multiple-objective combinatorial optimization",
  {\em Journal of Multi-Criteria Decision Analysis}, 7, 34--47.

Glover F. and Kochenberger, G.A. (2003), {\em Handbook of
Metaheuristics}, volume~57 of {\em International
  Series in Operations Research \& Management Science}, Boston, Dordrecht,
  London: Kluwer Academic Publishers.

Hansen, P. and Mladenovi\'{c}, N. (2003), ``Variable neighborhood
search", in Glover, F. and Kochenberger, G.A. (2003), chapter 6,
145--184.

Ishibuchi, H. and Murata, T. (1998), ``A multi-objective genetic
local search algorithm and its application
  to flowshop scheduling", {\em IEEE Transactions on Systems, Man, and Cybernetics},
  28, 392--403.

Johnson, D.S., Papadimitriou C.H. and Yannakakis, M. (1988), ``How
Easy Is Local Search?", {\em Journal of Computer and System
Sciences}, 37, 79--100.

Loukil, T., Teghem, J. and Tuyttens, D. (2005), ``Solving
multi-objective production scheduling problems using
  metaheuristics", {\em European Journal of Operational Research}, 161,
  42--61.

Louren\c{c}o, H.R., Martin, O. and St\"{u}tzle, T. (2003),
``Iterated local search", in Glover, F. and Kochenberger, G.A.
(2003), chapter 11, 321--353.

Murata, T., Ishibuchi, H. and Tanaka, H. (1996), ``Multi-objective
genetic algorithm and its application to flowshop
  scheduling", {\em Computers \& Industrial Engineering}, 30,
  957--968.

Nagar, A., Heragu, S.S. and Haddock, J. (1996), ``A combined
branch-and-bound and genetic algorithm based approach for
  a flowshop scheduling problem", {\em Annals of Operations Research}, 63,
  397--414.

Pinedo, M. (2002), {\em Scheduling: Theory, Algorithms, and
Systems}, Upper Saddle River, NJ: Precentice Hall.

Reeves, C.R. (1999), ``Landscapes, operators and heuristic search",
{\em Annals of Operations Research}, 86, 473--490.

Taillard, E. (1993), ``Benchmarks for basic scheduling problems",
{\em European Journal of Operational Research}, 64, 278--285.

T'kindt, V. and Billaut, J.-C. (2002), {\em Multicriteria
Scheduling: Theory, Models and Algorithms}, Berlin, Heidelberg, New
York: Springer Verlag.

Ulungu, E.L., Teghem, J., Fortemps, P.H. and Tuyttens, D. (1999),
``{MOSA} method: A tool for solving multiobjective combinatorial
  optimization problems", {\em Journal of Multi-Criteria Decision Making},
  8, 221--236.

Vincke, P. (1992), {\em Multicriteria Decision-Aid}, Chichester, New
York, Brisbane, Toronto, Singapore: John Wiley \& Sons.

%\bibliography{../../lit_bank,../../lit_bank_nv}
%\bibliographystyle{plain}

\end{document}